\documentclass{llncs}

\usepackage[utf8]{inputenc}
\usepackage[T1]{fontenc}

\usepackage[final]{graphicx}
\usepackage{epstopdf}
\usepackage{multirow}
\usepackage{tabu}
\usepackage{mathtools}
\usepackage{algorithm}
\usepackage[labelsep=period]{caption}
\usepackage[hyphens]{url}

\usepackage[russian,english]{babel}

\begin{document}

\title{Original Loop-closure Detection Algorithm for Monocular vSLAM}

\author{
Andrey Bokovoy\inst{1,3}
\and
Konstantin Yakovlev\inst{2,3}
}

\institute{
Peoples’ Friendship University of Russia (RUDN University), Moscow, Russia\\
\email{1042160097@rudn.university}
\and
Higher School of Economics, Moscow, Russia\\
\email{kyakovlev@hse.ru}
\and
Institute for Systems Analysis
of Federal Research Centre "Computer Science and Control"
of Russian Academy of Sciences, Moscow, Russia \\
\email{\{bokovoy,yakovlev\}@isa.ru}\\
}

\maketitle

\begin{abstract}
Vision-based simultaneous localization and mapping (vSLAM) is a well-established problem in mobile robotics and monocular vSLAM is one of the most challenging variations of that problem nowadays. In this work we study one of the core post-processing optimization mechanisms in vSLAM, e.g. loop-closure detection. We analyze the existing methods and propose original algorithm for loop-closure detection, which is suitable for dense, semi-dense and feature-based vSLAM methods. We evaluate the algorithm experimentally and show that it contribute to more accurate mapping while speeding up the monocular vSLAM pipeline to the extent the latter can be used in real-time for controlling small multi-rotor vehicle (drone).

\vspace{1em}
\end{abstract}

\section{Introduction}

Vision-based simultaneous localization and mapping (vSLAM) is one of the most challenging problems in computer vision and robotics. SLAM methods, that rely only on the information gained from minimum set of miniature passive sensors (monocular or stereo camera, inertial measurement unit), lie at the core of navigation capabilities of various mobile robots. Especially, they are of great value for compact unmanned aerial vehicles (which can not be equipped by the heavy, powerful sensors by default).

Recently a notable progress in the field of UAV vSLAM methods was made, see \cite{fu2014monocular,weiss2011monocular}, for example. However, there's still a large set of real-world problems and scenarios that can not be successfully tackled by the existing vision-based SLAM algorithms. The main reasons for that are the following.

First is the image processing time. Modern embedded computers that can be installed on compact UAVs are not that powerful to execute typical vSLAM pipelines in real time. Using external sources for remote computations is not always the solution since it lowers the mobility (robotic system is forced to continuously exchange huge amount of information with remote control station, using wire or wireless channel) and prevents robotic system from being fully autonomous. 

Second is poor image quality \cite{handa2014benchmark}. Small cameras typically mounted on compact UAVs are highly affected by the environment's conditions (light, weather etc.) and often produce video stream containing numerous jitters, noises and other artifacts. Thus one needs to apply different filtering techniques to pre-process the video stream and thus to improve the efficiency of vSLAM methods.

\begin{figure}[h]
  \centering
  \includegraphics[width=\textwidth]{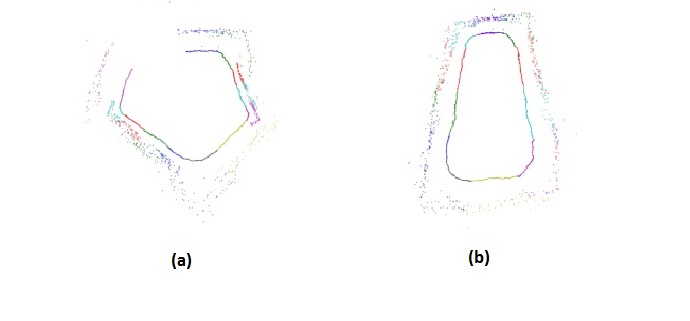}
  \caption{Solving SLAM problem with and without use of the loop-closure algorithm. (a) A raw map obtained with monocular vision-based SLAM method. The inner curve represents the trajectory of mobile robotic system. The outer points represent the map. (b) Trajectory and map optimized with the loop-closure algorithm}
  \label{fig:fig1}
\end{figure}

On top of that, all vSLAM methods are prone to accumulating error  \cite{strasdat2012visual} and that negatively affects the accuracy of constructed map and trajectory. One way to correct this error, and thus to increase the overall performance, is to handle, i.e. detect, \textit{loop-closures} - see fig.~\ref{fig:fig1}. More precisely one needs to detect that the current image comes from an already perceived scene and, in case it's true, correct the map and the trajectory.         

In this paper, we focus on improving the accuracy and performance of loop-closure detection algorithms. The ultimate goal is to keep the algorithm as robust and fast as possible along with making it compatible with  dense, semi-dense and feature-based vSLAM methods. We introduce two enhancement steps (within the loop-closure detection algorithm) that contribute towards reaching this goal.

The latter of the paper is organized as follows. In section 2, we present a brief overview of existing methods. Section 3 introduces our implementation of loop-closure algorithm. The experimental results, showing the accuracy and performance of implemented algorithm, are given in section 4. Section 5 concludes.

\section{Loop-closure methods}

Accumulating error is one of the main bottlenecks of almost all known monocular vSLAM methods and algorithms. Even state-of-the-art algorithms suffer from this  \cite{engel2016dso}. At the same time, results of numerous feasibility studies show, that detecting loop-closures can drastically improve the overall performance of monocular vSLAM. No wonder many of the vSLAM methods have loop-closure detection procedures built-in \cite{engel2014lsd,mur2015orb,mur2017visual}. There exist also standalone loop-closure detectors \cite{angeli2008fast,strasdat2011double} that may be plugged in to some of the vSLAM methods.

The earlier work \cite{botterill2011bag,konolige2010view,cummins2007probabilistic} mostly rely on the so-called global loop detection, when the current image was compared against  all previous visual data. This approach is quite reliable, but comes at the cost of high computation load and memory usage as one needs to keep all the information (such as keypoints, intense areas, depth map etc.) for every image processed during algorithm runtime. This leads to poor scaling for large environment localization and mapping. The recent approaches \cite{mur2015probabilistic,henry2014rgb,vokhmintsev2016simultaneous,buyval2015vision,afanasyev2015ros} use different constraints (e.i. using keyframes for keypoint matching) to optimize the time required for loop detection and map correction, but their usage is usually limited to specific vSLAM method.

In general loop-closure detection algorithms can be classified into three groups\cite{eade2008unified}: \textbf{map-to-map}, \textbf{image-to-map} and \textbf{image-to-image}:

\begin{itemize}
    \item \textbf{map-to-map} loop closure is done by splitting the global map into sub-maps and finding correspondences between them \cite{clemente2007mapping}. 
    \item \textbf{image-to-map} performs the search of the matches between image and a map and recovers the system's position, relative to the map \cite{williams2008image}.
    \item \textbf{image-to-image} founds a correspondences between images, usually based on vocabulary of image features \cite{cummins2008fab}.
\end{itemize}

Map-to-map approach is very intense performance-wise, since it deals with large amount of information on each iteration while comparing sub-maps. As the result it scales poorly to large environments. Image-to-map approach is fast and accurate, but in practice it is very memory intensive because one needs to store both point-cloud map and all the image features. The image-to-image loop-closure scales well to large environments, and can be computed fast with feature based approaches, but highly relies on a vocabulary. Thus one can infer that a combination of different approaches is desirable to reach higher performance while keeping the accuracy and the robustness at the high level. In this work we propose a solution that contributes towards this goal.

Proposed loop-closure method aims to combine image-to-image and image-to-map approach to achieve scalability, robustness  and accuracy of both approaches, while keeping moderate runtime and low memory usage. Besides the proposed method is compatible with a large number of existing vSLAM methods, including feature-based, semi-dense and dense vision-based SLAM methods (for monocular, stereo and RGB-D cameras) and can be seen as a general enhancement approach to loop-closure detection.


\section{Proposed method}
\begin{figure}[h]
  \centering
  \includegraphics[width=\textwidth]{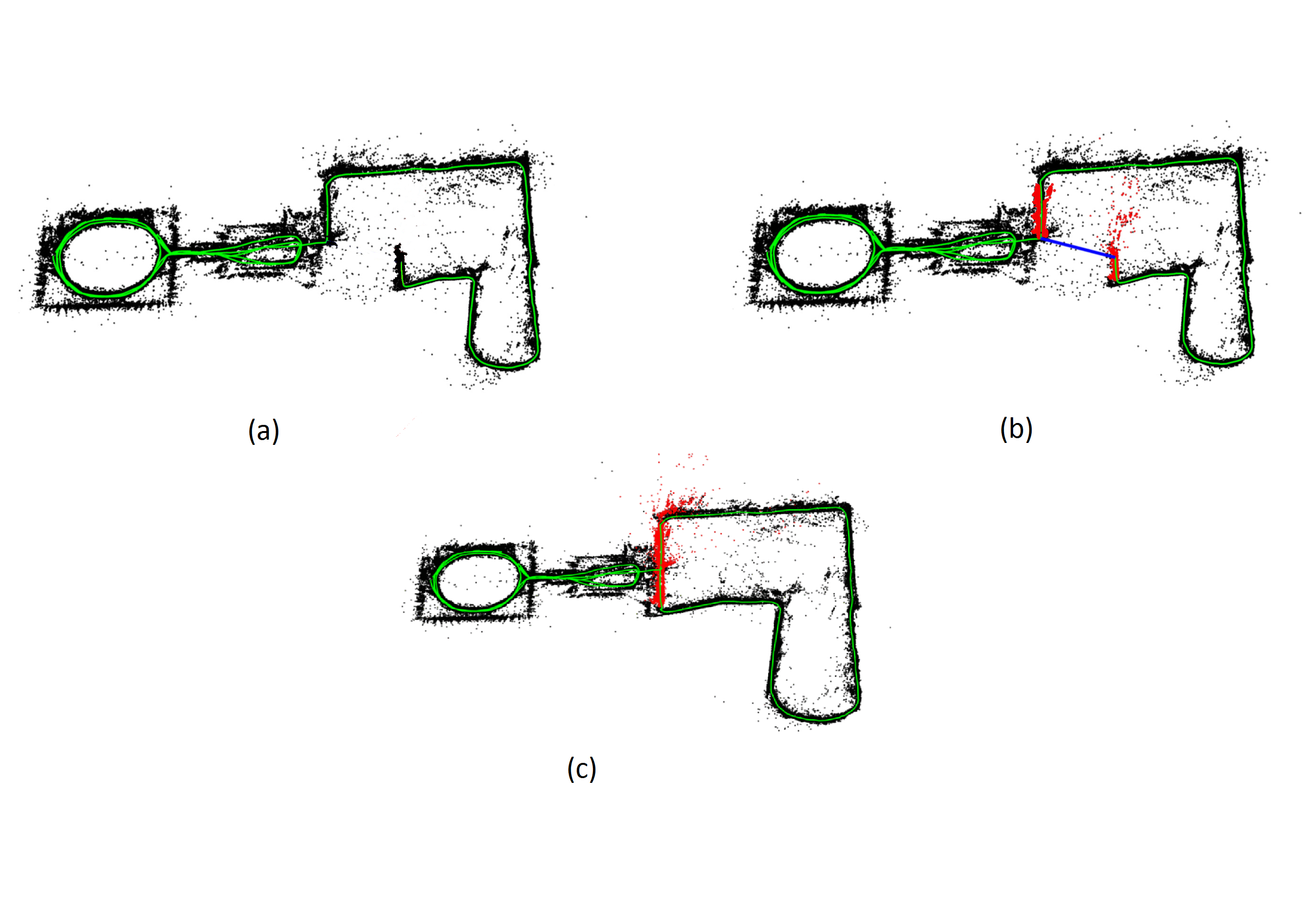}
  \caption{Main steps of the loop-closure detection. (a) Given (current) setting. Green curve represents the trajectory of the robotic system, black points represent the mapped features. (b) The loop-detection step. Red points represent the matched features, e.g. the ones that are present on the current image (perceived in current position) and previously observed images. (c) Map and trajectory of the robot after loop-closure detection.}
  \label{fig:fig2}
\end{figure}

In a nutshell all loop-closure algorithms generally consist of the two steps: 1) loop detection, 2) global optimization. Loop detection aims at establishing that the particular image is part of the scene, that has already been captured by previous image sequences. The simple interpretation is that this may be a sign, that the robotic system has reached the place that had already been visited before. The global optimization is performed after the loop is detected. This step corrects the accumulated run-time error for both the map and the trajectory (in a background). The illustration of the loop-detection process is depicted on fig.~\ref{fig:fig2}. 

Since the robotic system's motion consists of continuous rotations and translations, we assume that the trajectory is continuous as well (unless vSLAM method's tracking is lost), so loop-detection algorithm usually checks for trajectory loops once per $N$ images for performance optimization purposes. In cases, when tracking is lost, detection may be needed to recover the state and position of robotic system and rebuild the map.


We suggest 2 enhancement procedures to be performed while detecting the loop. They both aim at lowering down the number of features to be compared thus speeding up the algorithm. The enhancements include the image detection optimization and imposing geometric constraints. For fast and accurate image matching we found that storing a particular amount of informative keypoints (instead of all keaypoints) for each image allows us to keep the image matching accuracy. Also, the keypoints search area can be reduced to the only mapped points. The image comparison search area can be reduced by the geometric constrained, that is based on current camera position. We choose only images from the field, that may be observed from the camera in current position. High-level pseudocode of the loop-detection algorithm with the abovementioned procedures built-in is shown as Algorithm 1.

\begin{table}
\centering
\begin{tabular}{p{10cm}}
\hline
  \textbf{Algorithm 1} The proposed loop-closure detection algorithm. \\
\hline

\begin{enumerate}
\item Get an image from video flow
\item Extract keypoints and get their descriptors from corresponded mapped points on image
\item Get and store \textbf{K} informative features from image
\item \textbf{if} The trajectory loop is in camera search area
\item \quad Match corresponding images in search area with current image
\item \quad \textbf{if} the correspondence found
\item \quad \quad Perform the map optimization
\item \quad \textbf{endif}
\item \textbf{endif}
\end{enumerate} \\
 
\hline
\end{tabular}
\end{table}

First procedure (line 3) affects the feature extraction area of image. Dense and semi-dense vision-based SLAMs points of image with high gradient of intensity for depth map computation and mapping purposes. Thus, we reduce the extraction area by using only high gradient pixels, that were previously chosen to reconstruct the 3D space from 2D image. This allows us to avoid the image areas that are not going to be mapped anyway and provides an opportunity to reduce feature extraction process time. We limit the keypoints amount per any image in video flow to $K$. 
This $K$ keypoints with their descriptors are stored during loop-closure algorithm run-time since the number of keypoints per frame is relatively small (see section 4).


\begin{figure}[h]
  \centering
  \includegraphics[width=\textwidth]{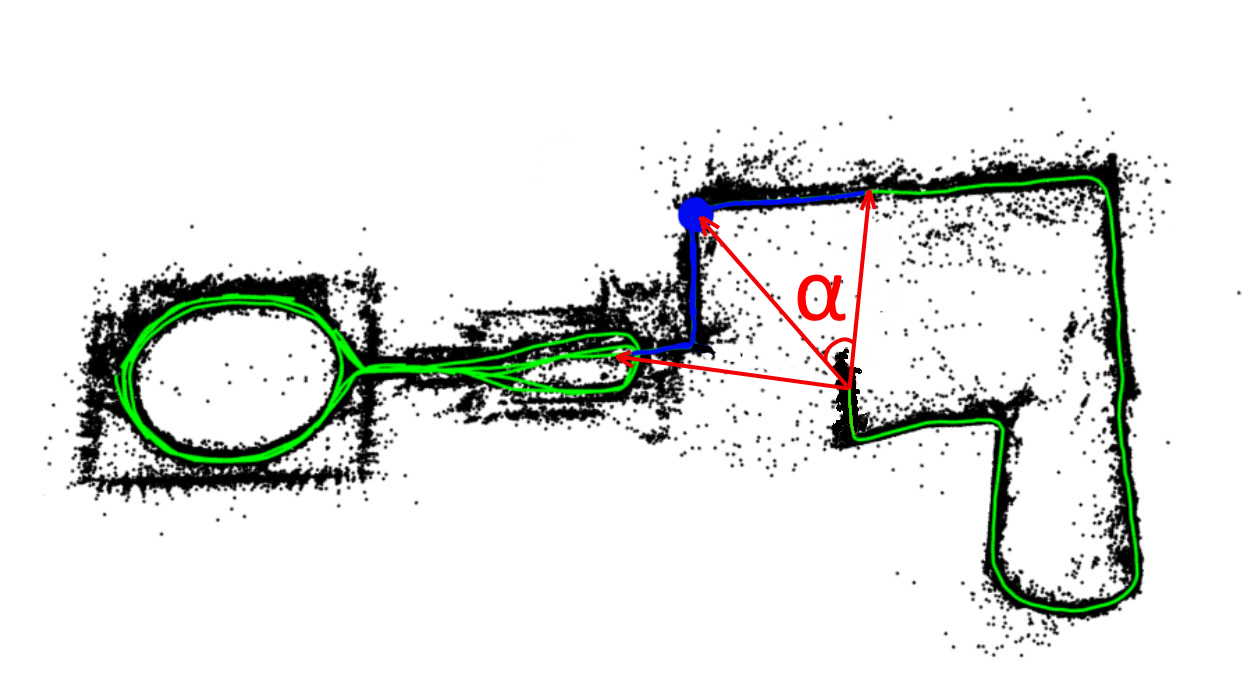}
  \caption{Loop detection search area. Red vectors shows the bounds for image comparison. Blue dot represents the start point for loop detection algorithm. And the trajectory picked out with blue color is our search area.}
  \label{fig:fig3}
\end{figure}

Second procedure (line 4) is the loop detection search area limitation. This allows to identify the patch on the whole trajectory that, with high probability, has a loop-closure point in it (i.e. the place, where the robot has already been). 
Assuming the robotic system's motion is mostly horizontal, we project the motion vector and continue it with a straight line. Then we draw a perpendicular to this line. If the perpendicular intersects the built trajectory, then we draw a $\alpha$ degree line between normal and the projected motion line. The closest position (with corresponding image) to the point of intersection is going to be a start point for loop detection algorithm with the whole loop detection area constrained by two points - the intersection of normal and motion line with trajectory.  

If motion line has no intersection point, then a search starts from the initial position of vision-based SLAM algorithm. The illustration of suggested method is demonstrated in fig.~\ref{fig:fig3}. 

More formally one can put it as follows. Assuming, that raw localization and mapping (without optimization) for each moment of time $t$ is done by vSLAM method. Thus, for a given moment of time $t$, we have a point cloud $M=\{m_i\}, i\in N$, that represents the map, sequence of images $I=\{I_1, I_2, ..., I_t\}$. For each image $I_t$ we have corresponding observation $z_t\in M$ and position vector $x_t$.     

As a part of loop-closure detection algorithm, we project each position $x_t$ on $xy$ plain $x'_t=Px_t$. For each $x'_k, k<t-1, k\in N$ we check if vector $\overrightarrow{w}=\overrightarrow{x'_{t-1} sx'_t}, s>0$ has intersection point $p$ with any of vectors $\overrightarrow{v}=\overrightarrow{x'_{k-1} x'_k}$ and vector $\overrightarrow{x'_1 (-lx'_2)}, l>0$. We assume the position $x'_k$ closest to intersection point $p$ to be the starting point for image matching. As an end point for image matching, we take the intersection point $p'$ of perpendicular $h$ to vector $\overrightarrow{x_t,x_{t-1}}$. As the result, the current image matching with corresponded images from positions between points $p$ and $p'$. 


\subsection{Implementation}

As the main image identifier for loop detection we've chosen ORB detector\cite{rublee2011orb} as one of the most fast, robust and efficient feature detector. For each image we extract at least $K$ ORB features and store their oriented and rotated BRIEF\cite{calonder2010brief} descriptors, that have high element sum, with associated images. The requirement of having element sum in BRIEF descriptors comes from their interpretation. Higher values mean higher intensity gradient at this points, that provides more robust feature matching. That means that such a keypoints are informative and can be stored for further image matching.  

As a part of map and trajectory global optimization, we use one of the most popular and effective graph optimization framework g$^2$o\cite{kummerle2011g}. That allows us to keep a high accuracy while optimizing map and trajectory in comparison to other modern vSLAM methods.

\section{Experimental results}

For performance and accuracy testing purposes of the developed method we use a Robot Operating System (ROS)\cite{quigley2009ros}, that provides a powerful tools for robotic algorithms researches in general and in vision-based SLAM testing in particular. The open-source realizations of ORB-SLAM and LSD-SLAM were taken as ones of the most popular feature-based and semi-dense SLAMs respectively. 

\begin{table}[h]
\caption{Loop detection success table}
\centering
\label{tab:representation}
\begin{tabular}{|c|c|c|c|c|c|c|c|c|c|c|c|c|}
\hline
\multicolumn{1}{|c|}{\multirow{2}{*}{\textbf{Dataset}}} & \multirow{2}{*}{\textbf{Method}} & \multicolumn{11}{c|}{\textbf{ORB Features}} \\ \cline{3-13} 
\multicolumn{1}{|c|}{} &  & \multicolumn{1}{c|}{\textbf{8}} & \multicolumn{1}{c|}{\textbf{9}} & \multicolumn{1}{c|}{\textbf{10}} & \multicolumn{1}{c|}{\textbf{11}} & \multicolumn{1}{c|}{\textbf{12}} & \multicolumn{1}{c|}{\textbf{13}} & \multicolumn{1}{c|}{\textbf{14}} & \multicolumn{1}{c|}{\textbf{15}} & \multicolumn{1}{c|}{\textbf{16}} & \multicolumn{1}{c|}{\textbf{17}} & \multicolumn{1}{c|}{\textbf{18}} \\ \hline
\multirow{2}{*}{Sequence 13} & ORB-SLAM & -- & -- & -- & -- & + & + & + & + & + & + & + \\ \cline{2-13} 
 & LSD-SLAM & -- & -- & -- & -- & -- & -- & + & + & + & + & + \\ \hline
\multirow{2}{*}{Sequence 14} & ORB-SLAM & -- & -- & + & + & + & + & + & + & + & + & + \\ \cline{2-13} 
 & LSD-SLAM & -- & -- & -- & + & + & + & + & + & + & + & + \\ \hline
\multirow{2}{*}{Sequence 15} & ORB-SLAM & -- & + & +  & + & + & + & + & + & + & + & + \\ \cline{2-13} 
 & LSD-SLAM & -- & + & + & + & + & + & + & + & + & + & + \\ \hline
\multirow{2}{*}{Machine} & ORB-SLAM & -- & -- & -- & -- & -- & -- & + & + & + & + & + \\ \cline{2-13} 
 & LSD-SLAM & -- & -- & -- & -- & -- & -- & -- & + & + & + & + \\ \hline
\multirow{2}{*}{Foodcourt} & ORB-SLAM & -- & -- & -- & -- & -- & + & + & + & + & + & + \\ \cline{2-13} 
 & LSD-SLAM & -- & -- & -- & -- & -- & -- & + & + & + & + & + \\ \hline
\end{tabular}
\end{table}

The introduced method is used with raw point cloud output of this methods. The experiment was made using LSD-SLAM Dataset\footnote{\url{http://vision.in.tum.de/research/vslam/lsdslam}}, KITTI vision benchmark suit\cite{geiger2013vision,Geiger2012CVPR}\footnote{\url{http://www.cvlibs.net/datasets/kitti/eval_odometry.php}} and Malaga Dataset\cite{blanco2014malaga}\footnote{\url{http://www.mrpt.org/MalagaUrbanDataset}}, which video fragments was divided into subsequences (distinguishing fragments with loops) to make the experimental research more relevant.

\begin{figure}    
\begin{minipage}[t]{0.45\textwidth}
\includegraphics[width=\linewidth]{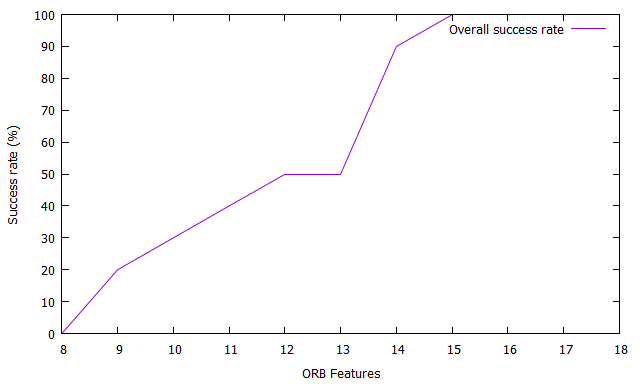}
\caption{Overall success rate of loop-closure detection algorithm with different amount of keypoints.}
\label{fig:fig4}
\end{minipage}
\hspace{\fill}
\begin{minipage}[t]{0.45\textwidth}
\includegraphics[width=\linewidth]{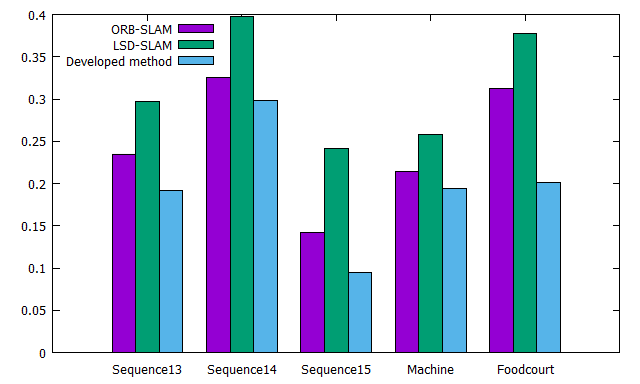}
\caption{The histogram shows run time (in seconds) for loop-closure algorithms used in ORB-SLAM and LSD-SLAM in comparison with our algorithm.}
\label{fig:fig5}
\end{minipage}

\begin{minipage}[t]{0.45\textwidth}
\includegraphics[width=\linewidth]{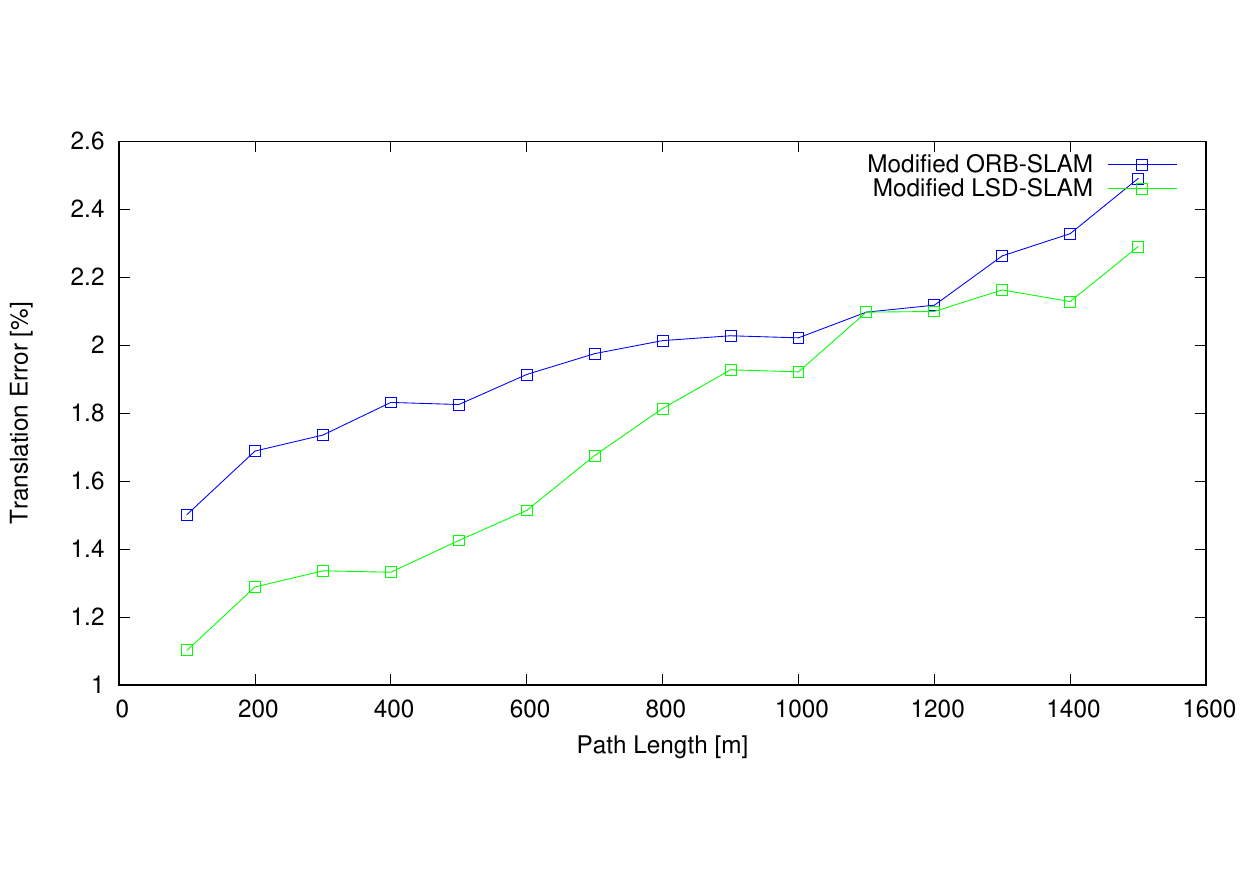}
\caption{Average translation error over 1.5 km distance for ORB-SLAM and LSD-SLAM with proposed loop-closure detection method.}
\label{fig:fig6}
\end{minipage}
\hspace{\fill}
\begin{minipage}[t]{0.45\textwidth}
\includegraphics[width=\linewidth]{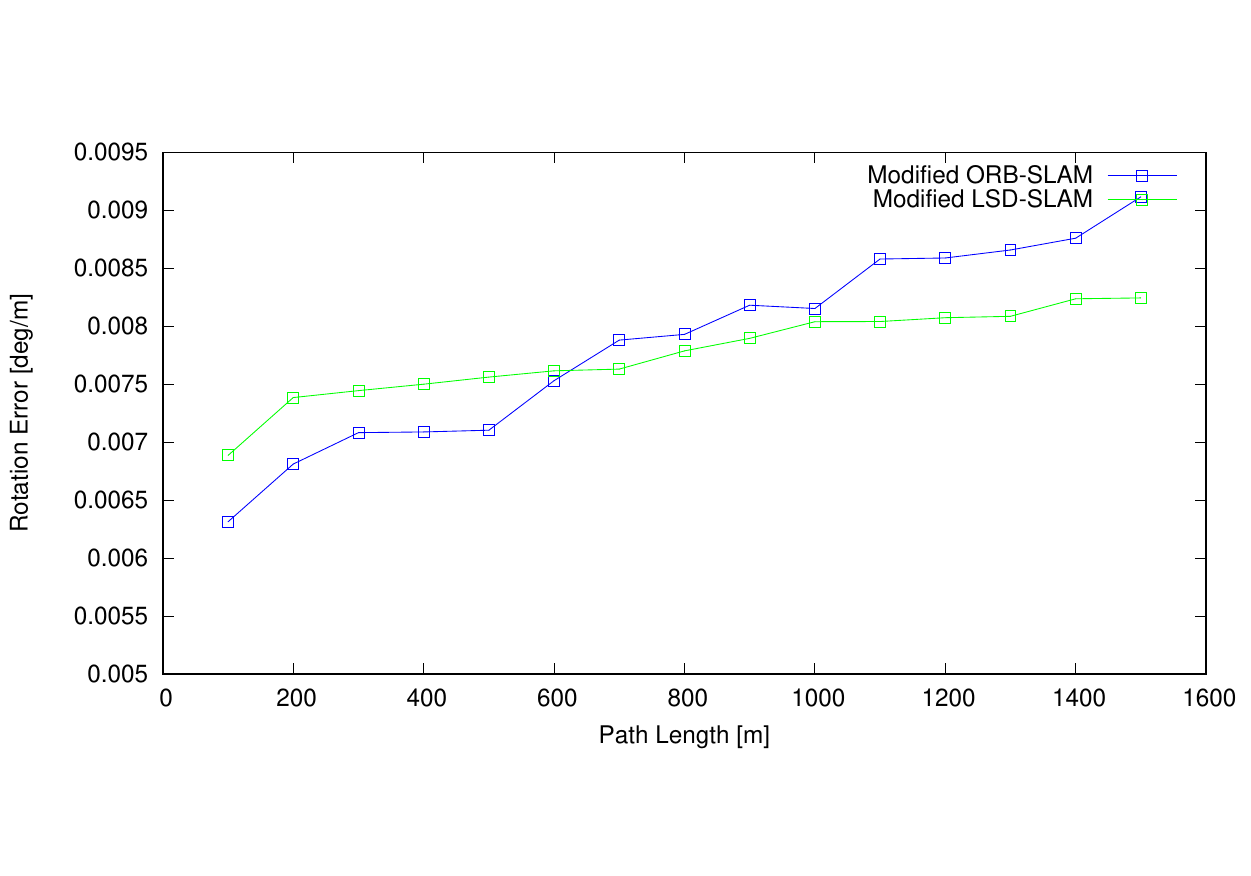}
\caption{Average rotation error over 1.5 km distance for ORB-SLAM and LSD-SLAM with proposed loop-closure detection method.}
\label{fig:fig7}
\end{minipage}

\end{figure}

We took the Sequences 13, 14 and 15 from KITTI dataset and Machine and Foodcourt Sequences from LSD-SLAM dataset, because that sequences contain trajectories with loop-closures. KITTI dataset includes ground truth, that allows us to compare the optimized trajectory with real one. For LSD-SLAM datasets we only test the performance of our algorithm and the accuracy in comparison with trajectories , built by LSD-SLAM and ORB-SLAM. For Malaga Dataset we took the whole 6th, 7th and 8th sequence, since they present single loop, and sequences 10 and 13 (which contain multiple loops) where divided into 7 and 3 single loop subsequences respectively. Thus, we used 13 sequences from Malaga Dataset.   

The first experiment was made to test the minimum required ORB features ($K$ value) for loop detection algorithm to function successfully (e.g. with 100\% success rate). The fig.~\ref{fig:fig4} shows the results of such an experiment. The table~\ref{tab:representation} shows the if the loop was successfully detected depending on number of ORB features ($R$) used for image matching.

\begin{figure}[h]
  \centering
  \includegraphics[width=\textwidth]{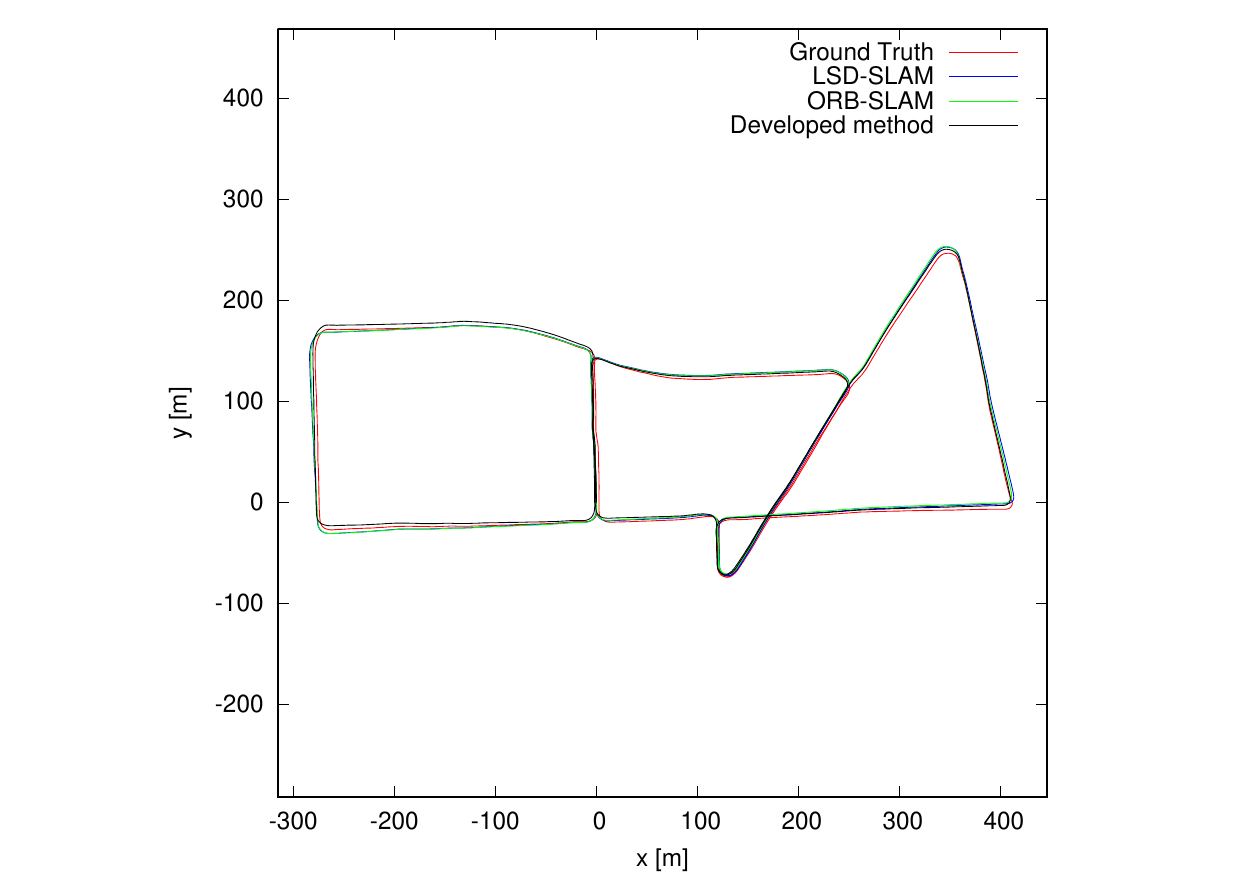}
  \caption{Ground truth comparison between LSD-SLAM, ORB-SLAM and developed method for Sequence 13.}
  \label{fig:fig8}
\end{figure}

As was already mentioned in section 2, we need to store at least $K$ features to successfully match the images if the loop-closure occurred. The experimental data shows that $K=15$ is a minimum value for loop to be detected. The presented result also allows us to dramatically reduce the memory usage, since we don't have to store hundreds of BRIEF descriptors, and increase the overall performance by the average of 7-10\% in comparison with LSD-SLAM's and ORB-SLAM's loop-closure algorithms as shown in fig.~\ref{fig:fig8}.

For KITTI and Malaga sequences, the trajectory ground truth is presented, so we tested our algorithm using the available data. Fig.~\ref{fig:fig6} shows the ground truth trajectory and the trajectory optimized with our method. The overall error values vary from 1.5\% to 2.5\% that is comparable with LSD-SLAM's and ORB-SLAM's loop-closure precision. 

The overall precision depends on trajectory's length and geometry. We found that longer trajectories with multiple loops give more accurate trajectory optimization for our method, while being more time consuming. 

\section{Conclusion}

We have developed the original loop-closure method, that can be used for dense, semi-dense and feature-based vSLAM methods. The introduced optimization techniques showed, that the combination of image-to-image approach for loop detection and image-to-map approach for global optimization keeps an accurate trajectory error correction (around 1.5-2.5\% translation error) while decreasing process time by 7-10\%.      

We found, that introduced method works in large outdoor environment without major issues. The experimental results showed, that described method can be used for mini unmanned aerial vehicle autonomous navigation tasks, even onboard.

\bigskip
\subsubsection*{Acknowledgment}
This research was supported by Russian Foundation for Basic Research. Grant 15-07-07483.

\bibliographystyle{splncs}
\bibliography{aistconf}

\end{document}